\documentclass[10pt,twocolumn,letterpaper]{article}

\usepackage{iccv}
\usepackage{times}
\usepackage{epsfig}
\usepackage{graphicx}
\usepackage{amsmath}
\usepackage{amssymb}
\usepackage{dblfloatfix}

\usepackage{makecell}
\usepackage{array}
\usepackage{multirow}

\usepackage[pagebackref=true,breaklinks=true,letterpaper=true,colorlinks,bookmarks=false]{hyperref}

\iccvfinalcopy 

\def\iccvPaperID{11211} 

\newcommand*{\subsubsubsection}[1]{\noindent\textbf{\vspace{-10pt}\newline#1.\hspace{1pt}}}

\ificcvfinal\fi

\begin{document}


\title{VTAMIQ: Transformers for Attention Modulated Image Quality Assessment}

\author{Andrei Chubarau \hspace{0.5cm} James Clark\\
Centre for Intelligent Machines, McGill University\\
Montreal, Canada\\
{\tt\small andrei.chubarau@mail.mcgill.ca \space clark@cim.mcgill.ca}
}

\maketitle

\ificcvfinal\thispagestyle{empty}\fi

\begin{abstract}

Following the major successes of self-attention and Transformers for image analysis, we investigate the use of such attention mechanisms in the context of Image Quality Assessment (IQA) and propose a novel full-reference IQA method, Vision Transformer for Attention Modulated Image Quality (VTAMIQ). Our method achieves competitive or state-of-the-art performance on the existing IQA datasets and significantly outperforms previous metrics in cross-database evaluations. Most patch-wise IQA methods treat each patch independently; this partially discards global information and limits the ability to model long-distance interactions. We avoid this problem altogether by employing a transformer to encode a sequence of patches as a single global representation, which by design considers interdependencies between patches. We rely on various attention mechanisms -- first with self-attention within the Transformer, and second with channel attention within our difference modulation network -- specifically to reveal and enhance the more salient features throughout our architecture. With large-scale pre-training for both classification and IQA tasks, VTAMIQ generalizes well to unseen sets of images and distortions, further demonstrating the strength of transformer-based networks for vision modelling.

\vspace{-6pt}


\end{abstract}
\section{Introduction}


As consumers are more and more accustomed to high fidelity imagery and video, image quality has become a critical evaluation metric in many image processing and computer vision applications. Directly asking human observers to rate visual quality requires slow and expensive subjective experiments. As a practical alternative, objective Image Quality Assessment (IQA) methods automate this process by mimicking the behaviour of the Human Visual System (HVS) for image quality evaluation. 

Given the highly subjective nature of human perception of image quality \cite{ssim2004, surveyIQA2011, tid2013}, naive computational IQA methods have limited success \cite{iqaIsHard}; more perceptually accurate full-reference (FR) image quality metrics (IQMs) require modeling key characteristics of the HVS, for instance its sensitivity to various visual stimuli based on luminance, contrast, chrominance, frequency content, etc. \cite{Barten1999ContrastSO, LeggeContrast1980, puEncoding2008}, as well as natural image statistics that describe image patterns and structures \cite{imageStats, naturalImageStatistics, bayesianIQA}. That being said, conventional IQA methods are often constrained by their respective assumptions about the HVS and are prone to underperform for practical IQA tasks \cite{iqaReviewBook}.


\begin{figure}[t]
\centering
\includegraphics[width=0.475\textwidth, trim={0.05cm 0.1cm 0.05cm 0.15cm}, clip] {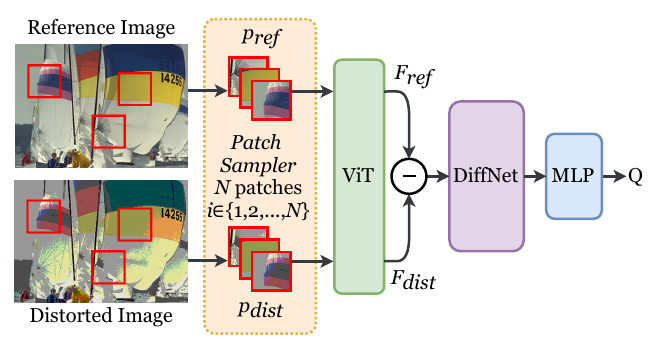}
\caption{
Simplified diagram of our proposed method, VTAMIQ. We apply context-aware probability sampling to extract sequences of $N$ spatially aligned fixed-size patches from the reference and the distorted images, and encode these using a Vision Transformer (ViT) \cite{vit2020} modified for sparse patch extraction. Our Difference Modulation Network (DiffNet) then recalibrates the resulting difference signal between the two encoded representations using a stack of residual channel-attention layers. Finally, a conventional fully-connected Multilayer Perceptron (MLP) regresses the modulated difference into a single image quality prediction.
\vspace{-10pt}
}
\label{fig:vtamiq_diagram}
\end{figure}

Following the major advances of deep Convolutional Neural Networks (CNNs) in many areas of computer vision research \cite{deeplearning}, the field of IQA has progressively gravitated towards deep learning \cite{cnnsIqa, deepQa, learningBlindIQA2013, deepNetworkIQA2016, NIMA2017, halucinatedIQA2018}. Jointly optimizing feature representations and inference directly from raw image data addresses the common limitations of classical approaches that rely on hand-crafted mechanisms. Granted, the available IQA datasets are typically modest in size and have limited image variety; this makes training deep IQMs from scratch a challenging task \cite{kadid10k, kadis700k, pieapp}. 

To facilitate training with limited data, many recent deep IQMs treat images as a combination of smaller-sized patches. Instead of tackling the full-resolution inputs directly, patch-wise metrics are independently computed and then aggregated for the final image quality prediction \cite{cnnsIqa, deepQa, WaDIQaM, pieapp, JNDSalcar}. An estimate of patch  importance or attention -- since attention essentially modulates the visibility of distortions \cite{vsi2014, saliencyIQA} -- can be used for weighted averaging of the patch-wise scores. Most existing methods, however, do not focus on complex attention models and simply process each patch independently. Patch scores and weights are estimated without complete global knowledge; the ability to account for large multi-scale features and long-distance interactions in the full-resolution input is thus limited.






Furthermore, arbitrarily defined signals in neural networks require specialized attention mechanisms. Convolutional operators only exploit local information within a predefined receptive field; their ability to model spatial and channel-wise dependencies is limited. Various attention strategies were thus introduced to complement conventional architectures with attention \cite{resnets, inception, hourglassAttention}. For instance, channel attention (CA) \cite{squeezeExcite, RCAN} and self-attention (SA) \cite{attentionIsAllYouNeed} explicitly evaluate the attention given to feature channels or elements of a sequence, respectively.

While CNNs dominate in computer vision, SA-based Transformers \cite{attentionIsAllYouNeed, transformerSurvey} have taken over natural language processing (NLP) due to their excellent ability to model sequences and long-distance dependencies, which the previous methods often lack \cite{longDistanceAttention, recurrent_attention, LSTM}. With enough data, appropriate pre-training, and clever architectural modifications, Transformers attain state-of-the-art results in vision tasks \cite{standAloneSelfAttention, vit2020} illustrating the flexibility and effectiveness of SA. 

Our key observation is the similarity between patch-wise IQA and transformers-based architectures: the input is a sequence of tokens or image patches. We further investigate this analogy along with various attention mechanisms in the context of IQA and propose a novel full-reference IQA method, Vision Transformer for Attention Modulated Image Quality (VTAMIQ). Our approach builds upon the success of Vision Transformer (ViT) \cite{vit2020} for vision tasks and leverages the ability of self-attention to jointly encode a sequence of image patches, by design considering patch interdependencies unlike previous patch-wise IQA methods. Employing a transformer allows VTAMIQ to avoid computing independent patch-wise quality metrics, and instead to directly determine a single latent representation for each input image, followed by a single global image quality score. A simplified diagram of our method is depicted in Figure \ref{fig:vtamiq_diagram}.

Furthermore, previous deep FR IQA methods can be typically broken down into several key steps: i) compute encoded representations (features) for the input images or patches, ii) compute difference between the encoded features, iii) correlate feature difference with image quality. Such approaches assume that the deep feature difference is immediately informative and can be directly correlated with image quality. We further propose an additional step: to intelligently modulate the difference between the two inputs such that more meaningful information is enhanced before regressing to a single quality score. We accomplish this by utilizing a specialized channel attention-based network inspired by the work in \cite{RCAN}.

Our method, VTAMIQ, achieves competitive or state-of-the-art IQA performance on the popular IQA datasets LIVE \cite{LIVEdatabase}, CSIQ \cite{CSIQ}, TID2013 \cite{tid2013}, and KADID-10k \cite{kadid10k}. We further emphasize the ability of our algorithm to generalize to unseen sets of images and distortions, demonstrated by VTAMIQ's vastly superior performance evaluated in cross-database tests. To further encourage the use of our metric as well as to facilitate reproducible research, our implementation and supplementary material will be made publicly available at \href{https://github.com/ch-andrei/VTAMIQ}{https://github.com/ch-andrei/VTAMIQ}.

    


\section{Related Work}
\label{sec:related_work}

A brief overview of IQA methods and relevant attention mechanisms is provided. We focus primarily on FR IQA as our method specifically falls under this category.

\subsubsubsection{Conventional IQA} Classical FR IQMs typically correlate image quality with some hand-crafted definition of perceptual difference between the inputs. The comparison can be based on error visibility \cite{nqm2000, vsnr2007}, structural similarity \cite{ssim2004, msssim2003, iwssim2011}, information content \cite{ifc2005, vif2006}, contrast visibility \cite{hdrvdp2005, hdrvdp2_2011}, or various other HVS-inspired feature similarities \cite{fsim2011, GMSD2013, mdsi2016, haarPSI2016, dog-ssim}. Combining IQA with visual saliency, i.e. models of human attention \cite{deepgaze}, further benefits prediction accuracy by targeting the more salient regions of the observed content \cite{vsi2014}.

\subsubsubsection{Deep Learning for IQA} To address the limitations of classical IQA approaches, data-driven deep learning methods determine a model of image quality that does not rely on hand-crafted definitions. While many of the existing IQA datasets are limited in size \cite{LIVEdatabase, CSIQ, tid2008, tid2013}, deep learning for IQA has become more and more feasible \cite{kadid10k, kadis700k}. Moreover, with the major successes of CNNs in many areas of computer vision research, exploiting commonalities between different tasks is possible with transfer learning \cite{transferLearning2011, transferLearning}. For instance, as FR IQA can be performed by comparing deep feature responses for two images \cite{cnnsIqa2016, lpips2018, deepFeaturesIQA2020}, it is possible to pre-train the underlying networks on large and widely available classification data \cite{imageNet, cifar}.

To further counter the lack of large IQA datasets, instead of directly training on full-resolution images, a common data augmentation approach is to explicitly subdivide the inputs into fixed-sized pixel patches. Patch-wise quality scores are then computed and aggregated into a single image quality prediction either by simple averaging \cite{cnnsIqa} or by a weighted pooling scheme using a form of visual sensitivity map \cite{deepQa}. For instance, WaDIQaM \cite{WaDIQaM} and PieAPP \cite{pieapp} directly estimate patch weights along with the patch quality scores. As an improvement, JND-Salcar \cite{JNDSalcar} further incorporates visual salience and just-noticeable-differences (JND) information for computing feature representations as well as for guiding patch-wise weight prediction.

Learning to rank -- and not directly score -- images based on quality is another possible interpretation of IQA. In \cite{RankIQA}, a network is trained to rank synthetically distorted images; the level of distortion is known, hence a ranking can be established. Ranking images can be further interpreted as the probability of preference of one image over another. Unlike in direct IQA measures, image quality is indirectly deduced from pairwise preference experiments. For instance, the work in \cite{pieapp} introduces a perceptual preference dataset PieAPP \cite{pieappWaterlooDataset} along with the PieAPP metric for IQA, a pairwise-learning framework for training an error-estimating function using preference probabilities. Perceptual error scores are computed for images $A$ and $B$, and the associated probability of preference is predicted; the trained error function can then be extracted as a baseline IQA model and further fine-tuned on the common IQA datasets.








\subsubsubsection{Channel Attention} In the context of deep feature responses with multiple channels (kernels), Squeeze-and-Excitation (SE) modules \cite{squeezeExcite, gatherExciteCnns} explicitly model interdependencies between channels to adaptively rescale the more salient channel-wise feature responses. Existing network architectures, for instance variants of VGG \cite{veryDeepCnns} and ResNet \cite{resnets}, can be enhanced with SE modules to outperform their original baselines. In \cite{RCAN}, SE is directly termed Channel Attention (CA); stacked CA modules are employed to build a deep residual network for super-resolution that specifically focuses on learning high-frequency information and significantly surpasses all previous approaches. 


\subsubsubsection{Self-Attention} A general attention function computes the relative importance of signals within a sequence. In self-attention (SA), the importance of each signal in the input is specifically estimated relative to the input sequence itself to compute its encoded representation. Conventional convolutional layers, by design, have a finite local receptive field; on the other hand, SA considers the entire input sequence and thus has a much stronger ability to model long-range interactions. While SA was first explored to favorably complement conventional CNNs \cite{attentionAugmentedCnns, longDistanceAttention}, it was later effectively applied in standalone applications averting the need for recurrence and convolutions \cite{attentionIsAllYouNeed, BERT}.


\subsubsubsection{Transformers} Whereas prior methods apply recurrent neural networks for sequence modelling \cite{longDistanceAttention, recurrent_attention, LSTM}, Transformers utilize an encoder-decoder architecture fully relying on self-attention to compute latent representations without the need for recurrent mechanisms or convolutions. Transformers scale well for large datasets and complex models, and can be easily parallelized, making up for their natural applicability for a wide variety of tasks \cite{transformerSurvey}. 

\subsubsubsection{Vision Transformer (ViT)} Inspired by the successes of Transformer-based architectures in NLP, the work in \cite{vit2020} applies a standard Transformer ``with the fewest possible modifications'' to vision tasks. Unlike NLP with word sequences, ViT operates on a sequence of flattened fixed-sized pixel patches extracted by tiling the input image. Although ViT only yields modest performance when trained directly on ImageNet \cite{imageNet}, with additional pre-training on large amounts of data and increased model complexity, ViT attains state-of-the-art classification performance at a fraction of training computational cost when compared to competitive methods.

\section{VTAMIQ: Vision Transformer for IQA}
\label{sec:VTAMIQ}

As illustrated in Figure \ref{fig:vtamiq_diagram}, our proposed FR IQA method, VTAMIQ, utilizes a modified Vision Transformer (ViT) to encode each input image as a single latent representation, computes the difference between encoded representations for the reference and the distorted images, intelligently modulates this difference, and finally, interprets the modulated difference as an image quality score.

While many previous deep FR IQA algorithms independently compute and aggregate patch-wise quality metrics, we directly avoid this by employing a Transformer to encode each input image as a single global representation. Although we still extract a sequence of spatially aligned patches to initially describe the reference and the distorted images, our model processes the input patches collectively and considers interdependencies between patches. More specifically, this is achieved with ViT computing self-attention with respect to the complete sequence of input patches. In this way, global information and long-distance interactions from the full-resolution image are taken into account and thereafter implicitly employed in the context of image quality assessment. 

The goal of our difference modulation network is then to enhance the more salient features of the difference between the encoded representations. Most deep IQMs directly use fully-connected layers to regress the difference vector; this assumes that the feature difference can be immediately correlated with image quality. Our results emphasize the importance of modulating the difference vector before applying regression: Channel Attention \cite{RCAN} (also known as Squeeze-and-Excitation \cite{squeezeExcite, gatherExciteCnns}) perfectly satisfies the requirements for this task.


\subsection{Feature Extraction}

\begin{figure}[!t]
  \centering
  \includegraphics[width=3in, trim={0 0.1cm 0 0.15cm}, clip] {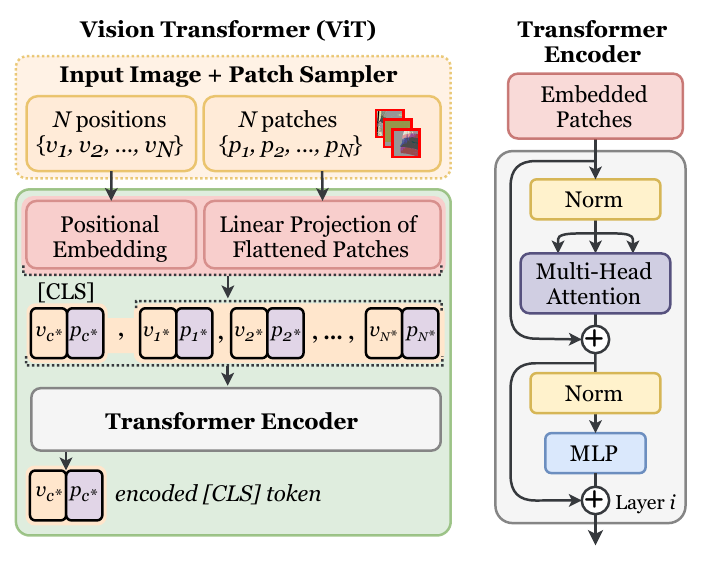}
  \caption
  {Vision Transformer (ViT) modified for VTAMIQ. Unlike the original ViT implementation where the raster order of tiled patches is relevant, we use patch center coordinates $v$ to index into the positional embeddings. After transforming the sequence, the encoded ``classification token'' \texttt{[CLS]} acts as the global representation of the input sequence. The architecture of the Transformer Encoder is shown on the right (figure adapted from \cite{vit2020, attentionIsAllYouNeed}).
  \vspace{-10pt}
  }
  \label{fig:vtamiq_vit}
\end{figure}

\begin{figure}[t]
  \centering
  \includegraphics[width=2.5in, trim={0.05cm 0.15cm 0 0.15cm}, clip] {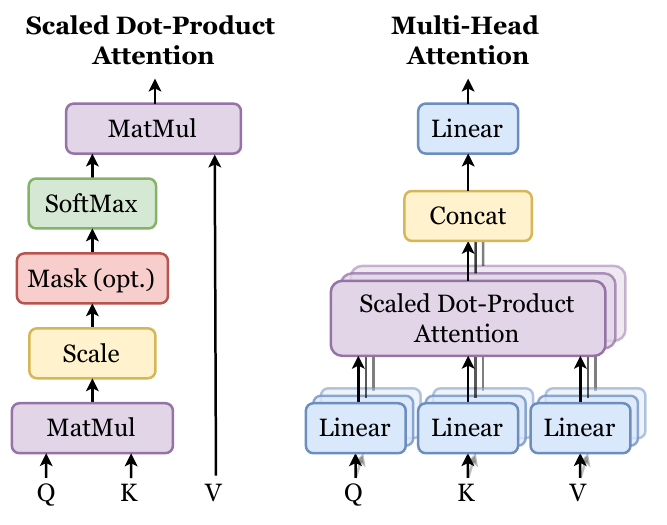}
  \caption{
  Scaled Dot-Product Attention (left) and Multi-Head Attention (right) used in Vision Transformer. Adapted from \cite{attentionIsAllYouNeed}.
  \vspace{-10pt}
  }
  \label{fig:multihead_attention}
\end{figure}

We use a Vision Transformer modified for sparse unordered patch input sequences to encode each input image as a vector of dimension $D$ corresponding to the hidden size of the transformer. As in ViT, the input image $I \in \mathbb{R}^{H \times W \times 3}$ is represented by $N$ fixed-sized RGB color patches $p_i \in \mathbb{R}^{N \times p \times p \times 3}$, where $p$ is the patch dimension. Patch embeddings are computed by applying a trainable linear projection to map from patch dimension to the latent Transformer dimension. Learned positional embeddings are then injected into the patch embeddings to account for positional information. The overall structure and components of ViT are illustrated in Figures \ref{fig:vtamiq_vit}-\ref{fig:multihead_attention}.

Although we reuse the overall structure of positional embeddings as in ViT, we employ a different strategy to assign positional embeddings to each patch. Instead of tiling the input image as in ViT, we use random sampling to extract $N$ patches; the resulting sequence is thus unordered and may contain overlapping patches. To address this, we use the positional $uv$ coordinate of each patch (available at the time of sampling), to index into the array of positional embeddings. This little modification complements the original ViT architecture allowing us to fine-tune a pre-trained ViT.

\subsection{Patch Sampling Strategies}

The simplest method to extract pixel patches is to tile an image. In the ideal scenario, we want to completely cover the full-resolution input, but this is often impractical due to memory and computational constraints associated with operating deep neural networks. Similarly to previous work, we instead opt to randomly sample a potentially sparse set of patches from the input image. This provides the hidden benefit of data augmentation which favors model training and reduces overfitting: each time an image is processed, a new and unseen random set of patches is extracted.

Previous patch-wise IQA methods employ naive tiling or random sampling to draw patch samples uniformly across the image; we improve on this by employing a special context-aware patch sampling strategy to extract more informative patches. We leverage key properties of human visual perception as well as the nature of full-reference IQA in our sampling scheme. First, to account for the centerbias of human visual attention \cite{deepgaze, saliencyBenchmarking}, we sample the more salient central regions with higher probability. Secondly, given the observation that extracting patches that are similar between the reference and the distorted images adds little information about image quality distortions, we bias the sampling towards regions with higher perceptual difference between the reference and the distorted images. An estimate of perceptual difference can be computed by simple metrics such as MSE or SSIM \cite{ssim2004} with similar results. 

Altogether, our context-aware patch sampling (CAPS) strategy allow us to use fewer patch samples for the same level of accuracy, effectively reducing memory and computational requirements. CAPS requires minimal overhead and speeds up model training, e.g. VTAMIQ can be trained about 15\% faster (fewer epochs) with CAPS when training with 256 patch samples per image. Note that the benefit of CAPS becomes less apparent when using larger patch counts as sampling variance is then naturally reduced.

\subsection{Deep Feature Difference Modulation}

\begin{figure}[t]
\vspace{-10pt}
  \centering
  \includegraphics[width=2.9in, trim={0.2cm 0.1cm 0cm 0.01cm}, clip] {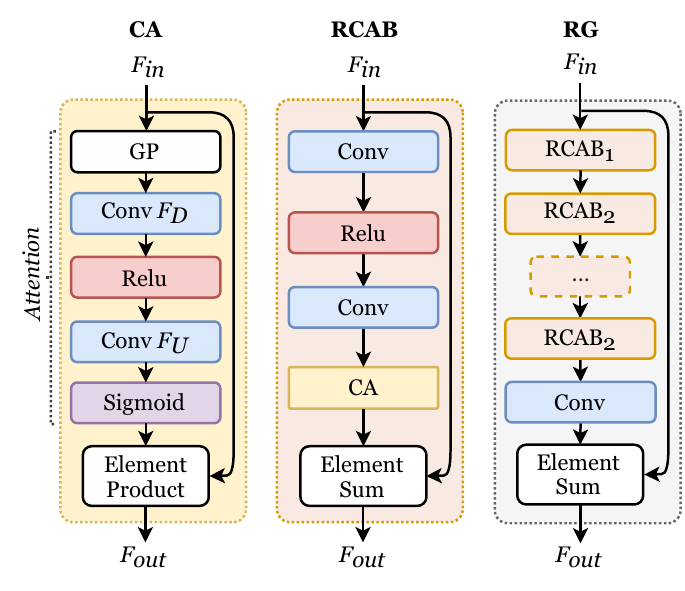}
  \caption
  {Components of VTAMIQ's Difference Modulation Network (DiffNet). Adapted from \cite{RCAN}.
  \vspace{-10pt}
  }
  \label{fig:rcab}
\end{figure}

We use Residual Channel Attention Blocks (RCABs) and Residual Groups (RG) proposed in \cite{RCAN} to modulate the difference between the encoded representations of the reference and the distorted images. Our Difference Modulation Network -- DiffNet for simplicity -- consists of a stack of $N$ RGs. Similarly, each RG consists of $M$ stacked RCABs, with the additional residual skip connection. Each RCAB further contains a skip connection linking the output of multiplicative Channel Attention (CA) and the input. The overall structures of CA, RCAB, and RG are depicted in Figure \ref{fig:rcab}. The underlying mathematical computations for DiffNet are defined in Equations \ref{eqn:diff_modulation_ca}-\ref{eqn:diff_modulation}, where $Attention$ corresponds to the learned ``excitation'' stage of CA responsible for computing channel weights \cite{squeezeExcite, RCAN}, and $U_{RG}$ and $U_{RCAB}$ are learnable linear transformations.

\setlength{\belowdisplayskip}{6pt}
\setlength{\abovedisplayskip}{-7pt}
\setlength{\jot}{8pt}
\begin{gather}
    CA(x) = x \times Attention(x) \label{eqn:diff_modulation_ca} \\
    RCAB(x) = x + U_{RCAB}(x)\times CA(x) \label{eqn:diff_modulation_rcab} \\
    RG(x) = x + U_{RG}(RCAB_N(...(RCAB_{1}(x))) \label{eqn:diff_modulation_rg} \\
    \mathit{DiffNet}(x) = RG_N(...(RG_{1}(x))) \label{eqn:diff_modulation}
\end{gather}

Previously in IQA, CA modules were used for feature extraction \cite{JNDSalcar} but never for modulating the difference vector. Moreover, in \cite{RCAN}, a construct similar to our DiffNet -- referred to as Residual in Residual (RIR) -- is shown to contribute to the extraction of more informative high-frequency features. Unlike the original RIR, we do not include the final residual skip connection in our variant to emphasize its signal rescaling properties. Lastly, unlike the original implementation with 10 RGs and a total of 200 RCABS \cite{RCAN}, our best IQA prediction accuracy was achieved with a shallower network with only 4 RGs each with 4 RCABs.

\subsection{Model Training}

Mean Absolute Error (MAE) and Mean Squared Error (MSE) are the commonly used loss functions for training IQA models \cite{cnnsIqa, deepQa, WaDIQaM, JNDSalcar}. Both functions compute the difference between the expected and the predicted values; due to the additional squaring operation, MSE penalizes larger differences more harshly than MAE. We evaluated training our model with MAE and MSE and found that more consistent results were obtained with MAE loss.


In addition to the above, we also found that optimizing with ranking loss \cite{RankIQA, lpips2018, JNDSalcar} benefits our model. Knowing that images can be ranked based on their quality -- e.g. image $A$ has higher quality than image $B$ -- ranking loss explicitly penalizes predictions that do not agree with the expected ordering. Ranking loss thus complements the more general MAE loss by offering a different guidance signal; it does not directly expose the magnitude of the expected predictions but specifically their relative ranking, which is arguably equally meaningfull in IQA. For a pair of image quality predictions, pairwise ranking loss can be computed as per Equation \ref{eqn:ranking_loss}, where $y_1$, $y_2$, $\hat{y}_1$, $\hat{y}_2$ correspond to the two predicted and the two expected scores, respectively, and $\epsilon$ is a small stabilizing constant. When the predicted ranking of the two images agrees with the expected result, the numerator value is negative, and hence the result is clamped to zero by the $max$ operator; when the ranking does not agree, ranking loss corresponds to the absolute difference between the two scores, thus pointing towards the correct ordering. 


\setlength{\belowdisplayskip}{4pt}
\setlength{\abovedisplayskip}{-8pt}
\setlength{\jot}{-2pt}
\begin{align} 
    L_{rank}(y_1, y_2, \hat{y}_1, \hat{y}_2) &= \label{eqn:ranking_loss} \\ \nonumber
    max & \left(0, \frac{-(\hat{y}_1 - \hat{y}_2) \times (y_1 - y_2) } {|\hat{y}_1 - \hat{y}_2| + \epsilon } \right)  
\end{align}

Naturally, in the context of training with mini-batches of size $N$, pairwise ranking loss can be computed for ${N \choose 2}$ possible pairwise combinations of the predictions; the total ranking loss then corresponds to the sum or the mean of the pairwise values. We experimented with combining MSE and ranking losses using different strategies and found that simple summation led to the optimal results.

\section{Experiments and Results}

\subsection{Datasets}

\begin{table}[b]
\vspace{-12pt}
\centering
\footnotesize
\setlength\tabcolsep{7pt}
\setlength\extrarowheight{-0.25pt}
\caption{Comparison of the available IQA datasets.}
\label{tab:datasets}
\begin{tabular}{c|ccc}
\Xhline{2\arrayrulewidth}\
\textbf{Dataset} & \textbf{Ref. images} &\textbf{Distortions} & \textbf{Dist. images} \\
\hline
LIVE \cite{LIVEdatabase} & 29 & 5 & 779 \\
CSIQ \cite{CSIQ} & 30 & 6 & 866 \\
TID2013 \cite{tid2013} & 25 & 24 & 3,000 \\ 
KADID-10k \cite{kadid10k} & 81 & 25 & 10125 \\
PieAPP \cite{pieapp} & 200 & 75 & 3,000 \\
KADIS-700k \cite{kadis700k} & 140,000 & 25* & 700,000 \\
\Xhline{2\arrayrulewidth}
\end{tabular}
\end{table}


We assess the performance of our method using the common major IQA datasets: LIVE \cite{LIVEdatabase}, CSIQ \cite{CSIQ}, TID2013 \cite{tid2013}, and KADID-10k \cite{kadid10k}. PieAPP \cite{pieapp} and KADIS-700k \cite{kadis700k} datasets are used for pre-training our model. Table \ref{tab:datasets} compares the listed datasets in more detail. Note that we omit the BAPPS dataset \cite{lpips2018} as it targets perceptual patch similarity and JND as opposed to image quality.

As recommended in \cite{kadid10k, criticalAnalysisIQA2021}, we randomly split the datasets into training ($\sim$60\%), validation ($\sim$20\%), and test sets ($\sim$20\%) along the reference image dimension. This ensures that the test and validation images are not seen by the network during training; validation set is then used for picking the best performing model, and the test set to evaluate the final performance. For LIVE, reference images are split into 17 training, 6 validation, and 6 test images; analogously for CSIQ, TID2013 and KADID-10k, the datasets are randomly split into 18-6-6, 15-5-5, and 49-16-16 subsets, respectively. PieAPP and KADIS-700k datasets are divided onto 160-40 and 120k-20k train-validation splits, respectively, and only used for pre-training our model.

\subsection{Pre-training a Baseline Model}
\label{sec:pretraining}

As reported in \cite{vit2020}, Vision Transformer strongly benefits from pre-training on larger datasets prior to transfer to down-stream tasks. To circumvent the lengthy requirements for training a Transformer from scratch, for all our experiments, we use ViT initialized with weights pre-trained on ImageNet and ImageNet-21k \cite{imageNet} classification databases\footnote{We did not implement or train ViT but re-used a publicly available pre-trained model from \href{https://github.com/jeonsworld/ViT-pytorch}{https://github.com/jeonsworld/ViT-pytorch} \cite{vitGithub}}. 

Our further pre-training for IQA consists of several steps. First, we pre-train VTAMIQ on the large KADIS-700k \cite{kadis700k} dataset, which contains 700,000 image pairs along with the corresponding scores given by 11 conventional IQMs as ``weak supervision'' signal for IQA. Out of the 11 available metrics, we select VSI \cite{vsi2014} scores as the prediction target. Unlike the conventional IQA datasets with limited number of reference images, KADIS-700k contains a wide variety of image data which benefits VTAMIQ's ability to generalize. Secondly, the resulting pre-trained model is fine-tuned on the subjective perceptual preference data from the PieAPP dataset \cite{pieapp} using the pairwise training framework proposed by the same authors. We find that this second pre-training stage serves as a further improvement for our baseline. We train for five epochs on each dataset and do not leave out a test set as no evaluations are performed. Note that as recommended in \cite{kadis700k}, we apply histogram equalization on the scores prior to training.

\subsection{Experimental Setup}

Since we use ViT models pre-trained on ImageNet \cite{imageNet} images, we normalize all input images with ImageNet's mean and variance from their original range. At training time, we extract 256 spatially aligned patches from the reference and the distorted images; for validation and testing, we use 1024 patches per input image to decrease the sparsity of sampling and improve the overall accuracy of image quality predictions. Patch sampling is redone each time an image is seen, thus even though the same source images are used, the extracted patches are different. We train with batch size set to 20 images, hence the pairwise rank loss is computed for 20 images in the batch. For evaluations, we run 20 epochs of training, where an epoch consists of seeing each image in the dataset once. For optimization, we use the AdamW optimizer \cite{adamw} with the recommended parameters and an initial learning rate (LR) of $10^{-5}$, decayed by a factor of 10 at epoch 12. Note that pre-training is done for 5 epochs: 3 with LR of $10^{-4}$, followed by 2 more at $10^{-5}$ LR. VTAMIQ converges to the optimal solution within the scheduled period as defined above. 

We implemented our proposed method VTAMIQ in Pytorch \cite{pytorch} and trained using a single NVIDIA GeForce RTX 3080 GPU with 10GB video memory. The practical training runtimes differ across datasets as the number of images in each dataset is different. On our setup and with the configurations defined above, training 20 epochs on KADID-10k dataset requires one hour, whereas a single epoch on KADIS-700k lasts 4 hours. 

\subsection{Model Configurations}

While exploring model design, we found that we may discard several top-most layers and only use the stem of ViT with little reduction to performance. This strategy is relatively common in transfer learning \cite{transferLearning}; specifically for layered architectures, earlier layers encode the general structure of images, whereas deeper layers typically extract more task-dependent information. Furthermore, we tested ViT with 32$\times$32 and 16$\times$16 patch sizes and determined that the latter exhibits superior IQA performance and generalization. As such, our best results are reported for VTAMIQ using a variant of ViT with 16$\times$16 patch size and the ``Base'' configuration (see \cite{vit2020}), keeping 6 out of the 12 available layers. We configure our DiffNet with 4 RGs each with 4 RCABS, and refer to this variant as VTAMIQ-16-6-4-4, a model with 57.3M parameters (43.6M from ViT).

\subsection{Performance Evaluation and Comparison}

\begin{table*}[t]
\centering
\footnotesize
\setlength\tabcolsep{4pt}
\setlength\extrarowheight{1pt}
\caption{Performance comparison on LIVE, CSIQ, TID2013, and KADID-10k Databases. Performance scores of other methods are as reported in the corresponding original papers. Best scores are \textbf{bolded}, second best are \underline{underlined}, missing scores are shown as ``--'' dash.}
\label{tab:performance_comparison}
\begin{tabular}{c|ccc|ccc|ccc|ccc}
\Xhline{2\arrayrulewidth}
  \multirow{2}{*}{\textbf{Method}}&
  \multicolumn{3}{c|}{\textbf{LIVE\cite{LIVEdatabase}}} &
  \multicolumn{3}{c|}{\textbf{CSIQ\cite{CSIQ}}} &
  \multicolumn{3}{c|}{\textbf{TID2013}\cite{tid2013}} &
  \multicolumn{3}{c}{\textbf{KADID-10k}\cite{kadid10k}} \\ 
  \cline{2-13}
  &
  \textbf{PLCC} &
  \textbf{SROCC} &
  \textbf{KROCC} &
  \textbf{PLCC} &
  \textbf{SROCC} &
  \textbf{KROCC} &
  \textbf{PLCC} &
  \textbf{SROCC} &
  \textbf{KROCC} &
  \textbf{PLCC} &
  \textbf{SROCC} &
  \textbf{KROCC} \\
  \Xhline{2\arrayrulewidth}
SSIM\cite{ssim2004}        & 0.942       & 0.948         & 0.812       & 0.861 & 0.876 & 0.691  & 0.691       & 0.637            & 0.464       & 0.723          & 0.724             & --              \\
MS--SSIM\cite{msssim2003}   & 0.946       & 0.956         & 0.829      & 0.899 & 0.913 & 0.739  & 0.833       & 0.786            & 0.608       & 0.801          & 0.802             & --              \\
FSIMc\cite{fsim2011}       & 0.953       & 0.963         & 0.846       & 0.919 & 0.931 & 0.769  & 0.877       & 0.851            & 0.667       & 0.851          & 0.854             & --              \\
MDSI\cite{mdsi2016}        & 0.966       & 0.967         & 0.840       & 0.953 & 0.957 & 0.813 & 0.909       & 0.890            & 0.712       & 0.873          & 0.872             & --              \\
VSI\cite{vsi2014}          & 0.946       & 0.948         & 0.807       & 0.928 & 0.942 & 0.786 & 0.900       & 0.897            & 0.718       & 0.878          & 0.879             & --              \\ 
SCQI\cite{scqi}            & 0.937       & 0.948         & 0.810       & 0.927 & 0.943 & 0.787 & 0.907       & 0.905            & 0.733       & 0.853          & 0.854             & --              \\ \hline
DOG--SSIMc\cite{dog-ssim}   & 0.966       & 0.963         & 0.844       & 0.943 & 0.954  & 0.813 & 0.934       & 0.926            & 0.768       & --              & --                 & --              \\
DeepFL--IQA\cite{kadid10k}  & 0.978       & 0.972         & --          & 0.946 & 0.930 & -- & 0.876       & 0.858            & --           & 0.938          & 0.936             & --              \\
DeepQA\cite{deepQa}        & 0.982       & 0.981         & --           & 0.965 & 0.961 & -- & 0.947       & 0.939            & --           & --              & --                 & --              \\
DualCNN\cite{dualCnn}      & --           & --             & --           & -- & -- & -- & 0.924       & 0.926            & 0.761       & 0.949          & 0.941             & 0.802          \\
WaDIQaM--FR\cite{WaDIQaM}   & 0.980       & 0.970         & --           & -- & -- & -- & 0.946       & 0.940            & 0.780       & --              & --                 & --              \\
PieAPP\cite{pieapp}        & \underline{0.986} & 0.977   & \underline{0.894} & 0.975 & 0.973 & \textbf{0.881}  & 0.946  & 0.945 & 0.804     & --              & --                 & --              \\
JND--SalCAR\cite{JNDSalcar} &
  \textbf{0.987} &
  \textbf{0.984} &
  \textbf{0.899} &
  \underline{0.977} &
  \underline{0.976} &
  0.868 &
  \textbf{0.956} &
  \underline{0.949} &
  \textbf{0.812} &
  \underline{0.960} &
  \underline{0.959} &
  -- \\ \hline
\textbf{VTAMIQ (ours)} & 
  0.984 & 
  \underline{0.982} & 
  0.891 & 
  \textbf{0.982} &
  \textbf{0.979} &
  \underline{0.875} &
  \underline{0.954} & 
  \textbf{0.950} & 
  \underline{0.809} & 
  \textbf{0.964} & 
  \textbf{0.963} & 
  \textbf{0.838} \\
\Xhline{2\arrayrulewidth}
\end{tabular}
\vspace{-3pt}
\end{table*}

\newcolumntype{C}{>{\centering\arraybackslash}m{1.5cm}}
\begin{table*}[t]
\centering
\footnotesize
\setlength\tabcolsep{6pt}
\setlength\extrarowheight{1pt}
\caption{
Performance comparison (SROCC) for cross-database evaluations. 
}
\label{tab:performance_crossdatabase_comparison}
\begin{tabular}{c|cc|cc|cc}
\Xhline{2\arrayrulewidth}
\textbf{Trained on:}  & \multicolumn{2}{c|}{\textbf{LIVE}}    & \multicolumn{2}{c|}{\textbf{TID2013}} & \multicolumn{2}{c}{\textbf{KADID-10k}} \\ \hline
\textbf{Tested on:}   & \textbf{TID2013} & \textbf{KADID-10k} & \textbf{LIVE}   & \textbf{KADID-10k}  & \textbf{LIVE}     & \textbf{TID2013}    \\ \hline
\Xhline{2\arrayrulewidth}
DOG-SSIMc \cite{dog-ssim}  & 0.751       & -     & \underline{0.948} & -     & -           & -           \\
WaDIQaM-FR \cite{WaDIQaM} & \underline{0.751} & -     & 0.936       & -     & -      & -      \\
DualCNN \cite{dualCnn}   & -           & -     & -           & -     & 0.751       & \underline{0.729} \\
DeepFL-IQA \cite{kadis700k} & 0.577       & 0.689 & 0.780       & 0.698 & \underline{0.894} & 0.71        \\ \hline
\textbf{VTAMIQ (ours)} & \textbf{0.874}   & \textbf{0.899}     & \textbf{0.957}  & \textbf{0.907}      & \textbf{0.974}    & \textbf{0.897} \\   
\Xhline{2\arrayrulewidth}
\end{tabular}
\vspace{-10pt}
\end{table*}

Performance of our model is assessed with the commonly used Pearson linear correlation coefficient (PLCC), Spearman rank order coefficient (SRCC), and Kendall rank-order correlation coefficient (KROCC). PLCC assesses the linear correlation between the expected and the predicted quality scores, whereas SROCC and KROCC describe the level of monotonic correlation between the two trends. The results are presented in Table \ref{tab:performance_comparison} where we report the mean of 20 evaluation runs. As with most recent IQA methods, we follow the suggestion in \cite{sheikhIqaFit} and apply non-linear fitting to our model's predictions prior to computing PLCC.

\newcommand*{\mcl}[1]{\multicolumn{1}{C|}{#1}}
\begin{table}[b]
\vspace{-10pt}
\centering
\footnotesize
\setlength\tabcolsep{2pt}
\setlength\extrarowheight{1pt}
\caption{Cross-database performance (SROCC) evaluation for VTAMIQ. Baseline is pre-trained on KADIS-700k and PieAPP.}
\label{tab:performance_crossdatabase}
\begin{tabular}{c|CCCC}
\Xhline{2\arrayrulewidth}
\multirow{3}{*}{\textbf{Trained on}} & \multicolumn{4}{c}{\textbf{Tested on}} \\ \cline{2-5} 
            & \mcl{\textbf{LIVE}} & \mcl{\textbf{CSIQ}}  & \mcl{\textbf{TID2013}}   & \textbf{KADID-10k} \\ \hline
\Xhline{2\arrayrulewidth}
Baseline  & \mcl{0.949}          & \mcl{0.936}           & \mcl{0.891}              & 0.895          \\ \hline
LIVE        & \mcl{\textbf{0.982}} & \mcl{0.955}  & \mcl{0.885}              & 0.899          \\
CSIQ        & \mcl{0.965} & \mcl{\textbf{0.979}}  & \mcl{0.893}              & 0.905          \\
TID2013     & \mcl{0.957}          & \mcl{0.947}           & \mcl{\textbf{0.950}}     & 0.907               \\
KADID-10k   & \mcl{0.974}          & \mcl{0.967}           & \mcl{0.897}              & \textbf{0.963} \\
\Xhline{2\arrayrulewidth}
\end{tabular}
\end{table}

\subsubsubsection{Comparison Against State-of-the-Art}
For all our tests, we fine-tune a baseline VTAMIQ pre-trained on KADIS-700k and PieAPP datasets. Our experiments demonstrate that VTAMIQ outperforms or is competitive with WaDIQaM \cite{WaDIQaM}, PieAPP \cite{pieapp}, and JND-SalCAR \cite{JNDSalcar} for all tested datasets. Especially on the larger KADID-10k database, we observe a solid improvement over previous work. Since VTAMIQ is built with ViT, it is a data-hungry Transformer-based model that improves with more training data. Besides, even conventional IQMs perform well on the smaller LIVE and CSIQ, but fall short on the more complex datasets such as TID2013 and KADID-10k. Furthermore, unlike JND-SalCAR \cite{JNDSalcar}, VTAMIQ does not require JND probability and salience maps and is directly applicable to the simple FR IQA problem involving a pair of images, without any additional inputs. As such, we both improve on the state-of-the-art, as well as reduce the input requirements of the previously best method. 

\subsubsubsection{Cross-Database Performance Evaluation} Due to limited training data as well as the high dimensionality of the input space, deep learning-based IQA methods tend to overfit on the datasets they are trained with, motivating the use of cross-database tests. We evaluate the cross-database performance of our method and demonstrate that VTAMIQ generalizes to unseen images and distortions more effectively than most previous methods. We compare VTAMIQ's cross-database performance to several prominent IQMs in Table \ref{tab:performance_crossdatabase_comparison}, clearly demonstrating the superior performance of our approach. Analogous improvement can be seen for all possible combinations of database cross-testing, with at least 10\% higher SROCC achieved by VTAMIQ. 

Full cross-database performance results of VTAMIQ are reported in Table \ref{tab:performance_crossdatabase}. Immediately after our pre-training stage (see Section \ref{sec:pretraining}), VTAMIQ already outperforms most conventional metrics. Furthermore, we observe that VTAMIQ does not ''forget'' its pre-training; its performance on unseen content only improves with more training data. For instance, a model pre-trained on KADIS-700k and LIVE performs better on the unseen TID2013 than a model pretrained on KADIS-700k alone. What is more, fine-tuning on the larger KADID-10k dataset offers more improvement than the smaller LIVE (despite LIVE and TID2013 partially sharing reference images). With that in mind, we foresee that the performance of our method can be further improved as more datasets are included in the training stage. Currently, we only train on FR IQA datasets; large NR IQA datasets (e.g. \cite{koniq10k, LiveitW}), can be leveraged in the future. 

\begin{table}[b]
\vspace{-10pt}
\centering
\small
\setlength\extrarowheight{1pt}
\caption{Performance (SROCC) comparison for models trained on PieAPP training set using the pairwise learning method from \cite{pieapp}. All evaluations use unseen test sets. *Results reported in \cite{pieapp}.}
\label{tab:pieapp}
\begin{tabular}{cccc}
\Xhline{2\arrayrulewidth}
\multicolumn{1}{c|}{\multirow{2}{*}{\textbf{Method}}} & \multicolumn{3}{c}{\textbf{Dataset}}                           \\ \cline{2-4} 
\multicolumn{1}{c|}{} &
  \multicolumn{1}{c|}{\textbf{\begin{tabular}[c]{@{}c@{}}PieAPP (test) \end{tabular}}} &
  \multicolumn{1}{c|}{\textbf{CSIQ}} &
  \textbf{TID2013} \\ \hline
\multicolumn{1}{c|}{DeepQA* \cite{deepQa}}                           & \multicolumn{1}{c|}{0.632} & \multicolumn{1}{c|}{0.873} & 0.837 \\
\multicolumn{1}{c|}{WaDIQaM-FR* \cite{WaDIQaM}}                       & \multicolumn{1}{c|}{0.748} & \multicolumn{1}{c|}{0.898} & 0.859 \\
\multicolumn{1}{c|}{PieAPP* \cite{pieapp}}        & \multicolumn{1}{c|}{\textbf{0.831}} & \multicolumn{1}{c|}{{\underline{0.907}}}    & {\underline{0.875}}    \\
\cline{1-4} 
\multicolumn{1}{c|}{\textbf{VTAMIQ (ours)}} & \multicolumn{1}{c|}{{\underline{0.829}}}    & \multicolumn{1}{c|}{\textbf{0.945}} & \textbf{0.908} \\
\Xhline{2\arrayrulewidth}
\end{tabular}
\end{table}

\subsubsubsection{Performance on Unseen Distortions} Recently with deep learning, some practical applications for IQA have shifted towards evaluating image restoration algorithms, e.g. deblurring, super-resolution, etc.; a lineup of ``novel'' generative distortions are now relevant. PieAPP dataset specifically contains a variety of such novel distortions and, unlike other IQA datasets, PieAPP's predefined test set purposefully holds out a disjoint set of distortions unseen during training. In Table \ref{tab:pieapp}, we assess VTAMIQ on the PieAPP test set to directly compare against PieAPP IQA metric and evaluate the ability of our method to generalize to unseen distortions. The results show that VTAMIQ is essentially on par with the PieAPP IQA metric but significantly outperforms all other previous work. That being said, the very same VTAMIQ model largely surpasses PieAPP metric in cross-dataset tests on unseen CSIQ and TID2013.


\subsection{Patch Sampling Strategies}

\begin{figure}[t]
  \centering
  \includegraphics[width=3.25in, trim={0.3cm 0.25cm 0.3cm 0.25cm}, clip] {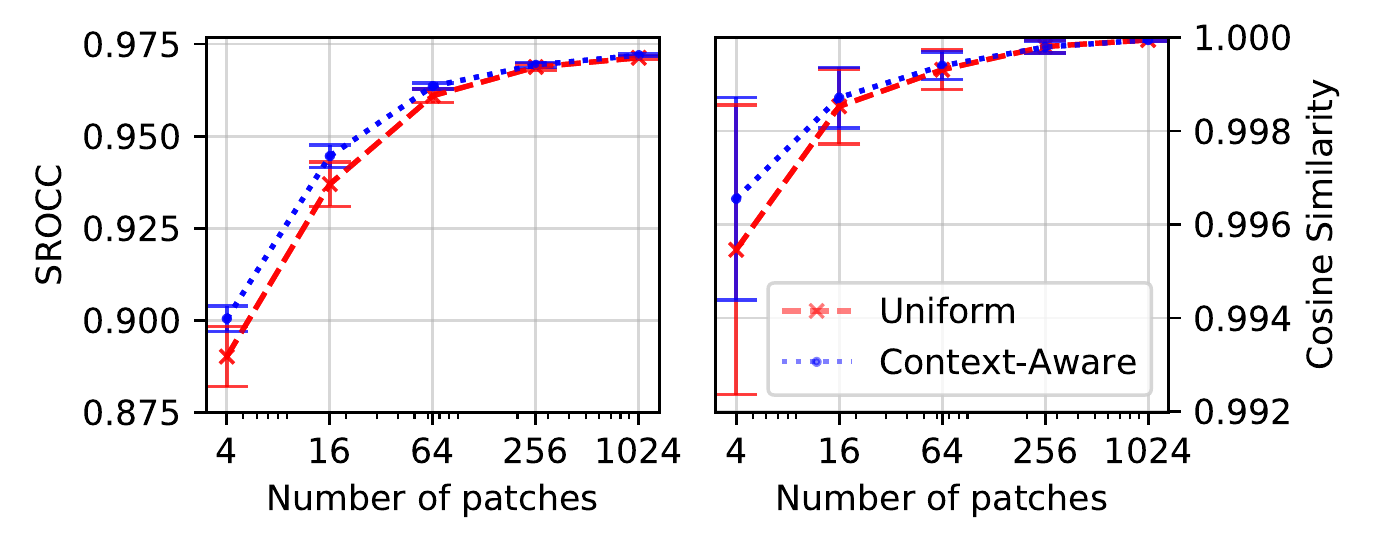}
  \caption
  {Influence of context-aware patch sampling and patch counts on VTAMIQ's performance on the LIVE dataset. On the left, the mean of SROCC for 10 runs is reported. On the right, we illustrate cosine similarity between the encoded representations for the same image acquired by resampling $N$ patches from the same image 128 times (averaged across all images in the LIVE dataset). Error bars denote 95\% confidence intervals.
  \vspace{-10pt}}
  \label{fig:live_patches_vs_srocc}
\end{figure}

We evaluate the effect of the number of patches and our context-aware patch sampling (CAPS) on VTAMIQ's performance and report the results in Figure \ref{fig:live_patches_vs_srocc} (using $16\times16$ patches and tested on the $512\times384$ images from LIVE dataset). Naturally, using more patches produces lower sampling variance and thus higher prediction accuracy. Especially for low sample counts, CAPS improves the consistency and the overall performance as shown by higher SROCC and smaller standard deviation values for the same number of extracted patches. That being said, using more than 512 patches results in increasingly minimal improvements. We further evaluate the cosine similarity between the encoded representations for images given randomized sequences of patches and similarly observe that CAPS improves the convergence of the encoded representations. As expected, the benefit of our sampling scheme becomes less apparent as the number of patches increases.


\subsection{Ablation Study}

\begin{figure}[!t]
  \centering
  \includegraphics[width=3.25in, trim={0.4cm 0.25cm 0.3cm 0.35cm}, clip]
  {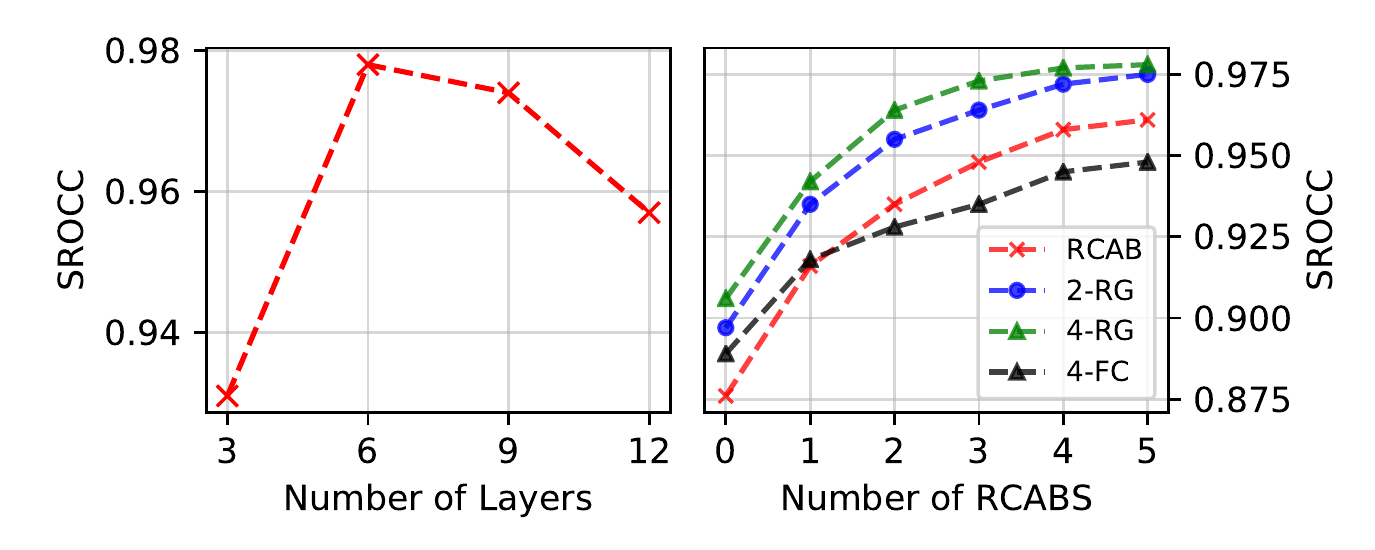}
  \caption
  {VTAMIQ's performance with different configurations for ViT and DiffNet as assessed on LIVE. On the left, we vary the number of retained layers in ViT. On the right, we evaluate DiffNet with varying number of Residual Groups (RG) and Residual Channel Attention Blocks (RCABs): i) no RGs and only RCABs (red), ii) two RGs (blue), and iii) four RGs (green), iv) four RGs with fully-connected layers instead of CA (black).
  \vspace{-10pt}
  }
  \label{fig:ablation_rcabs_rgs_vit}
\end{figure}

We perform ablations for VTAMIQ to validate our architectural choices, namely varying the configurations used for ViT and our Difference Modulation Network. The results are summarized in Figure \ref{fig:ablation_rcabs_rgs_vit}.


\subsubsubsection{ViT Configurations}
We investigate the influence of number of layers in ViT on VTAMIQ's performance and find that discarding several topmost layers results in very minimal performance loss. We determine that for the ``Base'' ViT configuration, keeping 6 layers in the Transformer is a good trade-off between model complexity and performance. Moreover, we observe that larger models are not as easily trained and often result in suboptimal IQA solutions despite having stronger performance for classification tasks. 

\subsubsubsection{Number of Residual Groups and RCABs}
VTAMIQ's quality estimation strongly relies on the employed difference modulation network. We ran ablations on the number of Residual Groups (RG) and the number of RCABs per RG in the modulation network. Judging by the results of our evaluation, the improvement of using a specialized difference modulating network is quite substantial: we find that without our modulation network, the performance deteriorates to that of early deep learning IQMs. We further determine that four RRs each with four RCABs offers a good balance between performance and model complexity. 

\subsubsubsection{Use of Channel Attention}
In order to directly validate the need for CA in our architecture, we replaced all CA modules with fully-connected layers (black plot in Figure \ref{fig:ablation_rcabs_rgs_vit}) and observe a significantly decreased SROCC on LIVE. We also validated VTAMIQ's performance with various configurations of RGs and RCABs (CA is part of RCAB): when using no RGs and zero RCABs -- and thus no CA -- VTAMIQ's performance is severely reduced. 

\section{Conclusion}

In this paper, we presented VTAMIQ, a Transformer-based difference modulation network for FR IQA that relies on various attention mechanisms to expose the more salient and informative features. We leverage a transformer to avoid the limitations of patch-wise IQA and instead operate with global feature representations. Our specialized difference modulation network then enhances the deep feature difference prior to predicting the final image quality score. With thorough pre-training, VTAMIQ outperforms the state-of-the-art on the common IQA datasets and robustly generalizes to unseen images and distortions with superior cross-database performance, further proving the effectiveness of attention mechanisms in vision modelling. 




{\small
\bibliographystyle{ieee_fullname}
\bibliography{vtamiq}
}

\end{document}